\documentclass[letterpaper]{article}

\usepackage[preprint, nonatbib]{neurips_2020}

\usepackage[utf8]{inputenc} % allow utf-8 input
\usepackage[T1]{fontenc}    % use 8-bit T1 fonts
\usepackage{hyperref}       % hyperlinks
\usepackage{url}            % simple URL typesetting
\usepackage{booktabs}       % professional-quality tables
\usepackage{amsfonts}       % blackboard math symbols
\usepackage{nicefrac}       % compact symbols for 1/2, etc.
\usepackage{microtype}      % microtypography
\usepackage{amssymb}
\usepackage{amsmath}
\usepackage{amsfonts}
\usepackage{algorithm}
\usepackage{algpseudocode}
\usepackage{bbm}
\usepackage{graphicx}
\usepackage{mwe}
\usepackage{caption}
\usepackage{subcaption}
\usepackage{wrapfig}

\newcommand{\kC}{\mathcal{C}}   %kaligraphisches C
\newcommand{\kM}{\mathcal{M}}   %kaligraphisches M
\newcommand{\kP}{\mathcal{P}}   %kaligraphisches P
\newcommand{\kX}{\mathcal{X}}   %kaligraphisches X
\newcommand{\Ind}{\mathbbm{1}} % Ind
\newcommand{\code}[1]{\texttt{#1}}

\newcommand\independent{\protect\mathpalette{\protect\independenT}{\perp}}
\def\independenT#1#2{\mathrel{\rlap{$#1#2$}\mkern2mu{#1#2}}}

\newtheorem{Def}{Definition}

\newtheorem{Satz}{Proposition}
\newenvironment{satz}[1]{\par {\bf Proposition #1} \par \itshape}{}

\newtheorem{Lemma}{Lemma}

\newenvironment{lemma}[1]{\par {\bf Lemma #1} \par \itshape}{}
\newenvironment{Proof}{\par {\it Proof.} \par}{\hspace*{\fill}
$\Box$\par\vskip2ex}    %Beweisumgebung

\newcommand{\DO}{{\rm do}}
\newcommand{\supp}{{\rm supp}}
\newcommand{\PN}{{\rm PN}}
\newcommand{\PS}{{\rm PS}}
\newcommand{\PNS}{{\rm PNS}}

\newcommand{\BIGOP}[1]{\mathop{\mathchoice%
{\raise-0.22em\hbox{\huge $#1$}}%
{\raise-0.05em\hbox{\Large $#1$}}{\hbox{\large $#1$}}{#1}}}
\newcommand{\bigtimes}{\BIGOP{\times}}

\title{Information-Theoretic Approximation to Causal Models}

\author{%
  Peter Gmeiner \\
  Global Data Science\\
  GfK SE, Germany;\\
  AlgoBalance UG (limited liability) \\
  \texttt{peter.gmeiner@gfk.com} \\
    \texttt{peter.gmeiner@algobalance.com} \\
}

\begin{document}

\maketitle

\begin{abstract}
Inferring the causal direction and causal effect between two discrete random variables $X$ and $Y$ from a finite sample is often a crucial problem and a challenging task. However, if we have access to observational and interventional data, it is possible to solve that task. If $X$ is causing $Y$, then it does not matter if we observe an effect in $Y$ by observing changes in $X$ or by intervening actively on $X$. This invariance principle creates a link between observational and interventional distributions in a higher dimensional probability space. We embed distributions that originate from samples of $X$ and $Y$ into that higher dimensional space such that the embedded distribution is closest to the distributions that follow the invariance principle, with respect to the relative entropy. This allows us to calculate the best information-theoretic approximation for a given empirical distribution, that follows an assumed underlying causal model. We show that this information-theoretic approximation to causal models (IACM) can be done by solving a linear optimization problem. In particular, by approximating the empirical distribution to a monotonic causal model, we can calculate probabilities of causation. We can also use IACM for causal discovery problems in the bivariate, discrete case. However, experimental results on labeled synthetic data from additive noise models show that our causal discovery approach is lagging behind state-of-the-art approaches because the invariance principle encodes only a necessary condition for causal relations. Nevertheless, for synthetic multiplicative noise data and real-world data, our approach can compete in some cases with alternative methods.
\end{abstract}

\section{Introduction}

Detecting causal relationships from data is a significant issue in many disciplines. The understanding of causal relations between variables can help to understand how a system behaves under intervention and has many other important implications. Identifying causal links (causal discovery) from observed data alone is only possible with further assumptions and/or additional data \cite{Pea09, SGS00, SHHK06}. Despite the various methods available, the problems of finding the causal structure and calculating the causal effect between two random variables remain notoriously hard. In this paper, we use additional data and assume a very natural principle to solve that task. Our work is based on {\em causal models}, which represent an (unknown) underlying data generation mechanism responsible for the distribution of the sampled data \cite{Pea09, PJS17}. We include sampled data from situations (environments) where interventions took place together with samples from pure observations. Recent developments in that direction revealed promising results \cite{PBM16, HPM18, ZHZGS17}, but often these methods are conservative, leading to situations where no direction is preferred. This paper focuses on the bivariate discrete case and is based on a natural and weak principle. The {\em principle of independent mechanism} assumes that the data generating mechanism is independent of the data that is fed into such a mechanism. From this principle, we derive an invariance relation that states that it does not matter if we observe an effect due to an observation of its cause or due to an intervention of its cause. Distributions that are generated by a causal model fulfil these invariance relations. If $X$ and $Y$ are discrete, then we can characterize the support set of joint distributions that fulfil these relations by embedding the distributions from observational and interventional samples into a higher dimensional space and creating a link between them. That means we first embed the empirical distributions into a higher dimensional space and then find the best approximation of this embedding to the probability distributions that are compatible with the invariance principle such that the relative entropy between them minimizes. We call this approach an information-theoretic approximation to causal models (IACM) since the relative entropy can be interpreted as an error telling us how much a finite sample deviates from a sample that comes from an assumed causal model. It turns out that solving this optimization problem is equivalent to solving a linear optimization problem. If we additionally assume in the binary case that the causal model is monotonic w.r.t. $X$ or $Y$ and include this assumption into the support set characterization used by our approach, we can calculate probabilities about how necessary, sufficient, or necessary and sufficient a cause is for an effect as defined in \cite{Pea09}. Furthermore, we formulate a causal discovery algorithm that uses IACM to infer the causal direction between two variables. For this, we approximate to a causal model where $X$ causes $Y$, and to a model were $Y$ causes $X$. We prefer the direction with lower relative entropy. With respective preprocessing, this can also be applied to continuous data.

For the rest of this paper, we consider two random variables $X$ and $Y$ with values in finite ranges $\kX_X$ and $\kX_Y$, respectively. The contribution of this paper is twofold. The first contribution is an approximation of distributions to a set of distributions that is compatible with an invariance condition induced by an assumed causal model. This enables us to calculate probabilities for causes in the binary case. The second contribution is a method for causal discovery based on this approximation procedure. This method can also be applied if we have observed data from $X$ and $Y$ that are heterogeneous and continuous. Compared to alternative state-of-the-art causal discovery methods our approach is not that strong because the invariance condition is only a necessary condition that a joint distribution originates from a causal model. Despite that weakness, in experiments with discrete real-world data, our approach could correctly reconstruct most of the causal directions. Compared to other information-theoretic methods for causal discovery, our proposal does not assume an explicit functional model like IGCI \cite{JMZLZDSS12} and ACID \cite{BV18} does. Our approach is a particular implementation of the Joint Causal Inference (JCI) Framework \cite{MMT20} with one context variable that formalizes perfect interventions and assumes JCI Assumptions 0-1. As for the invariant causal prediction (ICP) approach, IACM assumes that observational and interventional data are given explicitly. Otherwise, we use different preprocessing steps to split incoming data into such subsets.

The paper is organized as follows. Section \ref{sec.causalModels} introduces causal models and the invariance statement. In Section \ref{sec.iacm} we present an information-theoretic approximation of distributions to one that is generated by causal models. We derive the theoretic foundation, illustrate the results for the binary case, and formulate an approximation algorithm. Section \ref{sec.applications} shows the calculation of probabilities for causes as an application and a causal discovery approach based on our approximation. Section \ref{sec.experiments} describes experiments to evaluate our approach and we conclude in Section \ref{sec.discussion}.

\section{Causal Models} \label{sec.causalModels}

We describe causal relations in the form of a {\em directed graph} $G=(V,E)$ with a finite vertex set $V$ and a set of directed edges $E \subset V \times V$. A {\em directed edge} from $u \in V$ to $v \in V$ is an ordered pair $(u,v)$ and represented as an arrow between vertices, e.g. $u \rightarrow v$. For a directed edge $(u,v)$ the vertex $u$ is a {\em parent} of $v$ and $v$ is a {\em child} of $u$. The set of parents of a vertex $u$ is denoted by ${\rm PA}_u$. We only consider directed graphs without cycles and call them {\em directed acyclic graphs} (DAGs). In a DAG we interpret the vertices as random variables and a directed edge $(X_i, X_j)$ as a causal link between $X_i$ and $X_j$. We say that $X_i$ is a {\em direct cause} of $X_j$ and $X_j$ is a {\em direct effect} of $X_i$. 

\begin{Def}
A {\bf structural causal model} (SCM) is a tuple $\kC:=(S, P_N)$ where $S$ is a collection of $d$ structural assignments $X_j:=f_j ({\rm PA}_j, N_j), \; j=1, \ldots, d,$
where ${\rm PA}_j \subseteq \{X_1, \ldots, X_d\} \backslash \{X_j\}$ are the parents of $X_j$ and $P_N = P_{N_1, \ldots, N_d}$ is a joint distribution over the noise variables $N_j$ that are assumed to be jointly independent.
\end{Def}

We consider an SCM as a model for a data generating process \cite{Pea09, PJS17}. This enables us to model a system in an observational state and under perturbations at the same time. An SCM defines a unique distribution $P^{\kC}_X$ over the variables $X=(X_1, \ldots, X_d)$. Perfect {\em interventions} are done by replacing an assignment in an SCM. Given an SCM $\kC$ we can replace the assignment for $X_k$ by $X_k := \tilde f(\tilde{{\rm PA}}_k, \tilde N_k)$. The distribution of that new SCM $\tilde \kC$ is denoted by $P_X^{\tilde \kC} =: P_X^{\kC; \DO(X_k:=\tilde f(\tilde{{\rm PA}}_k, \tilde N_k))}$ and called {\em intervention distribution} \cite{Pea09, PJS17}. When modeling causality, we assume the {\em principle of independent mechanism}. Roughly speaking, this principle states that a change in a variable does not change the underlying causal mechanism, see \cite{PJS17}. Formally for an SCM, this means that a change in a child variable $X$ will not change the mechanism $f$ that is responsible to obtain an effect from $X$. From this principle the following invariance statement follows:

\begin{equation} \label{consistencyCondition}
p^\kC(x_j | x_{{\rm PA}_j}) = p^{\kC; \DO(X_k:=x)}(x_j|x_{{\rm PA}_j}),
\end{equation}
where $p^\kC(x_j|x_{{\rm PA}_j})$ is the conditional density of $P^\kC_{X_j|X_{{\rm PA}_j}=x_{{\rm PA}_j}}$ evaluated at $x_j$ for some $k \neq j$. Informally, this means that, if $X_k$ is a cause of $X_j$, then it doesn't matter if we observe $x_j$ when $x$ is present due to an observation of $X_k$ or when $x$ is present due to an intervention on $X_k$.

\section{Approximation to Causal Models} \label{sec.iacm}

\subsection{The General Case} \label{section.general_case}

Given two random variables $X, Y$ with finite ranges $\kX_X, \kX_Y$, and data from observations of $X, Y$ as well as from interventions on $X$ or $Y$.\footnote{We can also relax the assumption of having interventional data and assume that the data are heterogeneous and show a rich diversity. Alternatively, we can say that we have data of $X$ and $Y$ from different environments, where each environment belongs to a different intervention on $X$ or $Y$.} We further assume that the data from different interventions are independent of each other. In practical applications, the interventional data can be obtained from experiments or more implicitly from heterogeneous data. Condition (\ref{consistencyCondition}), is in general, not fulfilled by empirical distributions obtained from such data. We derive a method that enables us to find a joint probability distribution of $X$ and $Y$ that fulfil (\ref{consistencyCondition}) and is closest to an empirical distribution in an information-theoretic sense.

For the following derivation we assume the causal model $X \rightarrow Y$ with no confounding variables. Without loss of generality we assume that the intervention took place on $X$ with values in $\kX_X$. In the following we assume that $\kX_X$ is given as $\{1,\ldots, |\kX_X|\}$. We summarize ${\bf X}:= (X, (X_a)_{a \in \kX_X})$, ${\bf Y}:=(Y,(Y_a)_{a \in \kX_X})$, where $X,Y$ are the observed data and $(X_a)_{a \in \kX_X}, (Y_a)_{a \in \kX_X}$ the interventional data. We define $V:=\{X, Y\} \cup \bigcup_{a\in \kX_X} Y_{a}$ that takes values in $\kX_V:=\kX_X \times \kX_Y \times \bigtimes_{a \in \kX_X} \kX_{Y_{a}}$ and with $P_{V}$ we denote the joint distribution over $V$. The space of probability distributions on $\kX_V$ is denoted by $\kP(\kX_V)$ and for $A \subset V$ the marginalization of $P \in \kP(\kX_V)$ is defined by $\pi_A : \kP(\kX_V) \rightarrow \kP(\kX_A)$ with $\pi_A(P)(x):= \sum_{y\in \kX_{V \backslash A}} P(y,x)$, where $x\in \kX_A$ and $\kX_A := \bigtimes_{a \in A} \kX_a$. The next Lemma gives us a characterization of distributions that fulfil (\ref{consistencyCondition}).

\begin{Lemma} \label{lemma.characterization}
The set of joint probability distributions for $V$ which fulfil condition (\ref{consistencyCondition}) is given as
\begin{eqnarray*}
\kM_{X \rightarrow Y} & = & \left \{P \in \kP(\kX_V) \mid \pi_{X, Y, Y_{a}} P(a, y_a, \overline{y}_{a}) = 0 \right. \\
& & \qquad \qquad \qquad \left. \forall \; y_a \in \kX_{Y_a}, \overline{y}_a \in \kX_{Y_a} \backslash \{y_a\}, a \in \kX_X \right \}.
\end{eqnarray*}
\end{Lemma}
The proof is given in the Supplementary Material. The support of $\kM_{X \rightarrow Y}$ is given by 
\begin{eqnarray*}
\supp(\kM_{X \rightarrow Y}) = \! \!\!\!\!\!\!\!\!\!\!\!\! \bigcup_{a \in \{1,\ldots, |\kX_X|\}} \bigcup_{y \in \kX_Y} \{a\} \times \{y\} \times \kX_{Y_{x_1}} \times \ldots \times \kX_{Y_{a-1}} \times \{y\} \times \kX_{Y_{a+1}} \times \ldots \times \kX_{Y_{|\kX_X|}}.
\end{eqnarray*}
With observational and interventional samples of $X$ and $Y$ and its corresponding empirical distributions $P_{X,Y}$, $P_{Y_{a}}$ for $a \in \kX_X$ we try to find a distribution $\hat P \in \kM_{X \rightarrow Y}$ such that
\begin{equation} \label{equ.marginalization_problem}
\pi_{X,Y} \hat P = P_{X,Y}, \quad {\rm and}\; \pi_{Y_{a}} \hat P = P_{Y_{a}} \; {\rm for} \; a \in \kX_X.
\end{equation}
We can always find a joint distribution $\hat P \in \kP(\kX_V)$ such that (\ref{equ.marginalization_problem}) holds, since  the distributions $P_{XY}, P_{Y_{a}}$ for all $a \in \kX_X$ are independent to each other. Although this does not guarantee \mbox{$\hat P \in \kM_{X \rightarrow Y}$}, we can try to find a distribution in \mbox{$\kM_{X \rightarrow Y}$} that has minimal relative entropy to $\hat P$. This minimal relative entropy can be interpreted as an approximation error to the causal model $X \rightarrow Y$. The {\em relative entropy} or {\em Kullback-Leibler divergence} (KL-divergence) between two distributions $P, Q \in \kP(\kX_V)$ is defined as follows: \\
$
D(P || Q) := \left \{ \begin{array}{ll}
\sum_{x \in \kX_V} P(x) \log \left (\frac{P(x)}{Q(x)}\right ), & {\rm if\;} \supp(Q) \supseteq \supp(P), \\
\infty, & {\rm else.}
\end{array}
\right.
$ \\
We use the convention that $0 \log \frac{0}{q} = 0$ for $q > 0$, see also \cite{CT91, Kak99}. This leads to:

\begin{equation} \label{equ.MinMinProblem}
\min_{\substack{\hat P \in \kP(\kX_V),\\ \pi_{X,Y} \hat P = P_{XY}, \pi_{Y_{a}} \hat P = P_{Y_{a}},a\in \kX_X}} \min_{\tilde P \in \kM_{X \rightarrow Y}} D(\tilde P|| \hat P).
\end{equation}
That is a nonlinear min-min optimization problem with linear constraints. It turns out that in our situation, the problem simplifies to a linear optimization problem.

\begin{Satz} \label{prop.approx_solution}
The optimization problem (\ref{equ.MinMinProblem}) simplifies to the following linear optimization problem
$$
\max_{\substack{\hat P \in \kP(\kX_V),\\ \pi_{X,Y} \hat P = P_{XY}, \pi_{Y_{a}} \hat P = P_{Y_{a}},a\in \kX_X}} S_{X \rightarrow Y}(\hat P),
$$
with $S_{X \rightarrow Y}(\hat P):= \sum_{z \in \supp(\kM_{X \rightarrow Y})} \hat P(z)$.
\end{Satz}
The proof is given in the Supplementary Material and an application of the Lagrangian multiplier method. The statements of Proposition \ref{prop.approx_solution} holds also for any other support set characterization rather than $\kM_{X \rightarrow Y}$. The global approximation error is given by $D(\tilde P || \hat P) = -\log(S_{X \rightarrow Y}(\hat P))$. Our optimization procedure also minimizes a local approximation error $D(\pi_{X,Y} \tilde P || \pi_{X,Y} \hat P)$ that is smaller or equal than the global approximation error.

\subsection{The Binary Case} \label{sec.binary_case}

To illustrate our approach, we consider the binary case. That means $\kX_X = \kX_Y = \{0,1\}$, and $V=\{X,Y,Y_0, Y_1\}$. The set of consistent probability distributions is characterized by $$
\kM_{X \rightarrow Y}=\{P \in \kP(\kX_V) \;|\; P_{0010} = P_{0011} = P_{0100} = P_{0101} = P_{1001} = P_{1011} = P_{1100} = P_{1110} = 0\}
$$ 
and therefore $\supp(\kM_{X \rightarrow Y}) = \{0000, 0001, 0110, 0111, 1000, 1010, 1101, 1111 \}$. A probability distribution $\hat P \in \kP(\kX_V)$ is a non-negative vector with $16$ elements that sums up to $1$. We encode the conditions (\ref{equ.marginalization_problem}) into a contraint matrix $A$ 
%$$A =
%\begin{pmatrix}
%1 & 1 & 1 & 1 & 1 & 1 & 1 & 1 & 1 & 1 & 1 & 1 & 1 & 1 & 1 & 1 \\
%0 & 1 & 0 & 1 & 0 & 1 & 0 & 1 & 0 & 1 & 0 & 1 & 0 & 1 & 0 & 1 \\
%0 & 0 & 1 & 1 & 0 & 0 & 1 & 1 & 0 & 0 & 1 & 1 & 0 & 0 & 1 & 1 \\
%0 & 0 & 0 & 0 & 0 & 0 & 0 & 0 & 0 & 0 & 0 & 0 & 1 & 1 & 1 & 1 \\
%0 & 0 & 0 & 0 & 0 & 0 & 0 & 0 & 1 & 1 & 1 & 1 & 0 & 0 & 0 & 0 \\
%0 & 0 & 0 & 0 & 1 & 1 & 1 & 1 & 0 & 0 & 0 & 0 & 0 & 0 & 0 & 0 
%\end{pmatrix},
%$$
and a corresponding right-hand side $c$.
%\begin{eqnarray*}
%c & = &(1, P_{Y_0}(Y_0 = 1), P_{Y_1}(Y_1 = 1), P_{X,Y}(X = 1, Y=1), P_{X,Y}(X = 1, Y=0), \\
%& &\; P_{X,Y}(X = 0, Y=1)).
%\end{eqnarray*}
The non-negativity can be encoded in an identity matrix $\Ind_{16}$ of length $16$ and a zero vector $0_{16}$ of length $16$ as the right-hand side. A probability distribution $\hat P$ that solves (\ref{equ.marginalization_problem}) is a solution to the following linear problem
$$
\max S_{X \rightarrow Y}(\hat P) \; s.t. \; \; A \cdot \hat P = c \; {\rm and}\; \Ind_{16} \cdot \hat P \ge 0_{16}.
$$
The proof of Proposition \ref{prop.approx_solution} tells us that a distribution $\tilde P$ that fulfil condition (\ref{consistencyCondition}) and is as close as possible to $\hat P$ in an information-theoretic sense can be obtained by the following re-weighting of $\hat P$
$$
\tilde P (x) := \frac{\hat P(x)}{S_{X \rightarrow Y}(\hat P)}, \quad {\rm if}\; x \in \supp(\kM_{X \rightarrow Y}) \quad {\rm and}\quad \tilde P(x) := 0, \quad {\rm if}\; x \notin \supp(\kM_{X \rightarrow Y}).
$$

\subsection{Implementation}

The procedure in Subsection \ref{sec.binary_case} can be generalized for arbitrary finite ranges and more complex causal models. The pseudo-code of the algorithm is shown in Algorithm \ref{alg.approximationAlgo}. The size of the finite ranges is denoted by $b_x := |\kX_X|$ and $b_y := |\kX_Y|$. We further assume that we have for every $x \in \kX_X$ interventional data available. Therefore, the constraint matrix $A$ has dimension $b_x (2b_y -1) \times b_x b_y^{b_x +1}$. The first row of $A$ contains $1$ at each column, the following $b_x(b_y-1)$ rows contain the support patterns of $P_{Y_{a}}$ and the final $b_x b_y -1$ rows contain the support pattern of $P_{XY}$. The function \code{getConstraintDistribution} prepares the right-hand side of $A$ accordingly. The algorithm accepts a joint distribution $P$ of $X,Y$ and a causal model $C$ (for example $X \rightarrow Y$) as parameters and is formulated for more complex causal models, see Subsection \ref{subsec.extened_models}. We implemented this procedure in Python and used the \code{cvxpy} package to solve the linear program.\footnote{The code for the presented algorithms is available at \url{https://github.com/pgmeiner/iacm}.} The dimension of $A$ will grow exponentially with the size of ranges for $X$ and $Y$. However, it turns out that it is enough to consider low range sizes $b_x \le 4, b_y \le 4$. For possibly preprocessed sample data of $X$ and $Y$ with higher or continuous range sizes, we apply an equal-frequency discretization based on quantiles, that is done before calculating and feeding $P$ into Algorithm \ref{alg.approximationAlgo}.
\begin{algorithm}
\caption{IACM($P$, $b_x$, $b_y$, $C$)}
\label{alg.approximationAlgo}
\begin{algorithmic}[1]
\State $A \gets {\rm \code{createConstraintMatrix}}(b_x, b_y)$
\State $c \gets {\rm \code{getConstraintDistribution}}(P, b_x, b_y)$
\State Solve LP problem: $\max S_{C}(\hat P)$ s.t. $A \hat P = c$ and $\Ind_{b_x b_y b_y^{b_x}} \hat P \ge 0_{b_x b_y b_y^{b_x}}$
\State $\tilde P(x) \gets \left \{ \begin{array}{ll} \frac{\hat P(x)}{S_{C}(\hat P)}, & \;{\rm for}\; x\in \supp(\kM_{C}), \\
0, & \; {\rm for}\; x\notin\supp(\kM_{C}). \end{array} \right.$
\State $D_{C} \gets -\log(S_C(\hat P))$ or $D_{C} \gets D(\pi_{X,Y} \tilde P|| \pi_{X,Y} \hat P)$ depending on the setting
\State return $\tilde P, D_{C}$
\end{algorithmic}
\end{algorithm}

\subsection{Extension to other causal models} \label{subsec.extened_models}
The procedure described above can be applied in the same manner to other causal models whenever we have a support set characterization for condition (\ref{consistencyCondition}). If we consider $X$ as a family of variables on which interventions took place and $Y$ as a family of variables that have been observed, then Algorithm \ref{alg.approximationAlgo} generalizes to more complex causal models, see Supplementary Material.

\section{Applications} \label{sec.applications}

\subsection{Probabilities for Causes}

To measure the effect of a cause-effect relation in a causal model $X\rightarrow Y$ Pearl proposed in Chapter 9 of \cite{Pea09} {\em counterfactual statements} that give information about the necessity, the sufficiency, and the necessity and sufficiency of cause-effect relations. A {\em counterfactual statement} is a do-statement in a hypothetical situation that can, in general, not observed or simulated. Formally this means we condition an SCM to an observed situation and apply a do-operator. The corresponding intervention distribution reads for example $P^{\kC |(X,Y)=(1,1);\DO{(X=0)}}(Y=0)$ which means the probability that $Y$ equals $0$ if $X$ would have been $0$ where indeed we observed that $X$ is $1$ and $Y$ is $1$.

\begin{Def}
Let $X, Y$ be random variables in an SCM $\kC$ such that $X$ is a (hypothetical) cause of $Y$ and $x \in \kX_X, y \in \kX_Y$.
\begin{itemize}
	\item The probability that $X=x$ is necessary as a cause for an effect $Y=y$ is defined as
	$\PN_{x\rightarrow y} := P^{\kC|(X,Y)=(x,y);\DO(X\in \overline{x})}(Y\in \overline{y}),$
	where $\overline{x} = \kX_X \backslash \{x\}$.
	\item The probability that $X=x$ is sufficient as a cause for an effect $Y=y$ is defined as
	$\PS_{x\rightarrow y} := P^{\kC|(X,Y)\in(\overline{x},\overline{y});\DO(X=x)}(Y=y).$
	\item The probability that $X=x$ is necessary and sufficient as a cause for an effect $Y=y$ is defined as $\PNS_{x\rightarrow y} := P(X=x,Y=y) \PN_{x\rightarrow y} + P(X\in \overline{x},Y\in \overline{y})\PS_{x\rightarrow y}.$
\end{itemize}
\end{Def}

In general, counterfactual statements cannot be calculated from observational data and without knowing the true underlying SCM. However, Pearl identified situations in which we can exploit the presence of observational and interventional data to calculate the probabilities defined above. One such situation is when the underlying SCM is monotonic.

\begin{Def}
An SCM $\kC$ with $Y:= f(X, N_Y)$ with $X \independent N_Y$ is called {\em monotonic} relative to $X$, if and only if $f$ is monotonic in $X$ independent of $N_Y$.
\end{Def}

If $X$ and $Y$ are binary and if $Y$ is increasing monotonic relative to $X$, then Theorem 9.2.15 in \cite{Pea09} give us
\begin{equation} 
\begin{aligned} \label{eq.PN_PS_PNS_incr}
\PN_{1\rightarrow 1} & = \frac{P(Y=1) - P^{\kC; \DO(x=0)}(Y=1)}{P(Y=1, X=1)},\quad \PS_{1\rightarrow 1} = \frac{P^{\kC; \DO(x=1)}(Y=1) - P(Y=1)}{P(Y=0,X=0)}, \\
\PNS_{1\rightarrow 1} & = P^{\kC; \DO(x=1)}(Y=1) - P^{\kC; \DO(x=0)}(Y=1).
\end{aligned}
\end{equation}
Similar, if $Y$ is decreasing monotonic relative to $X$, then we could also derive in the same fashion as Pearl did it the following formulas
\begin{equation} 
\begin{aligned} \label{eq.PN_PS_PNS_decr}
\PN_{0\rightarrow 1} &= \frac{P^{\kC; \DO(x=1)}(Y=0)- P(Y=0)}{P(Y=1, X=0)}, \quad \PS_{0\rightarrow 1} = \frac{P(Y=0)-P^{\kC; \DO(x=0)}(Y=0)}{P(Y=0,X=1)}, \\
\PNS_{0\rightarrow 1} &= P^{\kC; \DO(x=0)}(Y=1) - P^{\kC; \DO(x=1)}(Y=1).
\end{aligned}
\end{equation}
By approximating empirical observational and interventional distributions to a monotonic causal model we can calculate $\PN_{x\rightarrow y}$, $\PS_{x\rightarrow y}$, and $\PNS_{x\rightarrow y}$. For this we need to further restrict the set $\kM_{X \rightarrow Y}$ and note that the monotonicity of $f$ implies that either $\{Y_0=1, Y_1=0\}$ has zero probability or that $\{Y_0=0, Y_1=1\}$ has zero probability. This means that either $P_{0110} = P_{1010} = 0$ or $P_{0001} = P_{1101}=0$ has to hold in addition to the conditions given in $\kM_{X \rightarrow Y}$. We define $\kM_{{X \rightarrow Y}, M_i} := \{P \in \kM_{X \rightarrow Y} | P_{0110} = P_{1010} = 0\}$ as the set of probability distributions with an underlying monotonic increasing data generation process and $\kM_{{X \rightarrow Y}, M_d} := \{P \in \kM_{X \rightarrow Y} | P_{0001} = P_{1101} = 0\}$ as the set of probability distributions with an underlying monotonic decreasing data generation process.
An approximation in the sense of Subsection \ref{section.general_case} to those monotonic models will only change the definition of $S_{X \rightarrow Y}(P)$, the rest will remain the same. In order to calculate $\PN_{x\rightarrow y}$, $\PS_{x\rightarrow y}$, and $\PNS_{x\rightarrow y}$ we approximate to $\kM_{{X \rightarrow Y},M_i}$ and $\kM_{{X \rightarrow Y}, M_d}$, choose the one with the least approximation error and use the formulas given above. We state the pseudo-code in Algorithm \ref{alg.CalcCausalProbabilities}.

\begin{algorithm}[h]
\caption{CalcCausalProbabilities($P$)}
\label{alg.CalcCausalProbabilities}
\begin{algorithmic}[1]
\State $\tilde P_{i},  D_{i} \gets {\rm IACM}(P, 2, 2, ({X \rightarrow Y},M_i))$
\State $\tilde P_{d},  D_{d} \gets {\rm IACM}(P, 2, 2, ({X \rightarrow Y},M_d))$
\If {$D_{i} < D_{d}$} 
	\State Calculate $\PN, \PS, \PNS$ using $\tilde P_i$ and formulas (\ref{eq.PN_PS_PNS_incr})
\Else
	\State Calculate $\PN, \PS, \PNS$ using $\tilde P_d$ and formulas (\ref{eq.PN_PS_PNS_decr})
\EndIf
\State return $\PN$, $\PS$, $\PNS$
\end{algorithmic}
\end{algorithm}

\subsection{Causal Discovery}

In general a distribution $P \in \kM_{X \rightarrow Y}$ need not to come from a data generation mechanism that follows a causal model $X \rightarrow Y$ (for example the uniform distribution over $\kX_V$ is also in $\kM_{X \rightarrow Y}$). This is because condition (\ref{consistencyCondition}) is only necessary but not sufficient for a causal relation, and therefore we cannot expect identification results for causal relations only based on $\kM_{X \rightarrow Y}$.\footnote{However, we can further restrict $\kM_{X \rightarrow Y}$ for binary $X,Y$ based on the findings of Section III C. in \cite{PJS11}, that lead to more stable results. See Supplementary Material for more details.} Despite this fact, it is worth to consider IACM for causal discovery. With IACM we can test how well given datasets fit to an assumed causal model $C$ and obtain a metric $D_{C}$ that quantifies this. We can use this to identify the causal direction between $X$ and $Y$. The direction with the smallest metric is the one we infer as the causal direction. If the difference between the metrics is below a small tolerance $\epsilon>0$, we consider both directions as equal and return "no decision". If $X$ and $Y$ are binary and the error to monotone models is smaller than to non-monotone models, then we apply Algorithm \ref{alg.CalcCausalProbabilities} to determine $\PNS$ for both directions and use this as a decision criterion for the preferred direction (the direction with the higher $\PNS$ is the preferred one). In general, some kind of data preprocessing and discretization before applying the causal discovery method is of advantage. In our implementation, we include several different preprocessing steps that try to split the input data into observational and interventional data w.r.t. $X$ or $Y$ if such data not given explicitly. See Supplementary Material for more details. Algorithm \ref{alg.iacmdiscovery} shows the pseudo-code of our causal discovery approach.

\begin{algorithm}
\caption{IACMDiscovery(X, Y, $b_x$, $b_y$)}
\label{alg.iacmdiscovery}
\begin{algorithmic}[1]
\State ${\rm data}_X \gets$ preprocessing of $X, Y$ w.r.t. $X$
\State ${\rm data}_Y \gets$ preprocessing of $X, Y$ w.r.t. $Y$
\If {$b_x = b_y = 2$ AND monotone model is preferred}
	\State use CalcCausalProbabilities to get ${\rm PNS}, D_{X \rightarrow Y}, D_{Y \rightarrow X}$ for $X\rightarrow Y$ and $Y \rightarrow X$
	\State If $|D_{X \rightarrow Y} - D_{Y \rightarrow X}|<\epsilon$, then return direction with highest ${\rm PNS}$
\Else
	\State $D_{X \rightarrow Y} \gets {\rm IACM}(P_{{\rm data}_X}, b_x, b_y, \kM_{X \rightarrow Y})$
	\State $D_{Y \rightarrow X} \gets {\rm IACM}(P_{{\rm data}_Y}, b_x, b_y, \kM_{Y \rightarrow X})$
	\State If $|D_{X \rightarrow Y} - D_{Y \rightarrow X}|<\epsilon$, then return no decision
\EndIf
\State If $D_{X \rightarrow Y} < D_{Y \rightarrow X}$, then return $X\rightarrow Y$  else return $Y\rightarrow X$
\end{algorithmic}
\end{algorithm}

\section{Experiments} \label{sec.experiments}

We test Algorithm \ref{alg.iacmdiscovery} with synthetic and real-world data against alternative causal discovery methods. IACM runs with a preprocessing procedure described in the Supplementary Material and uses $D(\pi_{X,Y} \tilde P||\pi_{X,Y} \hat P)$ as an approximation error, which is more sensitive than the global approximation error. We also tested a version of IACM with additional constraints (called IACM+) based on findings of Section III C. in \cite{PJS11} that is explained in the Supplementary Material.

\subsection{Pairwise Causal Discovery Methods}

Among the various causal discovery approaches for continuous, discrete, nonlinear bivariate data, we select those that do not include any training of labeled cause-effect pairs to have a fair comparison. One well-known method uses additive noise models that assume SCMs with additive noise and applies for continuous and discrete data \cite{HJMPS09, PJS11}. We select the version for discrete data (DR). Furthermore, we select an information-geometric approach (IGCI) \cite{JMZLZDSS12} designed for continuous data and some recent methods for discrete data that are using minimal description length (CISC) \cite{BV17}, Shannon entropy (ACID) \cite{BV18}, and a compact representation of the causal mechanism (HCR) \cite{CQZZH18}. We further select regression error based causal inference (RECI) \cite{BJWSS18}, and invariant causal prediction (ICP) \cite{PBM16} as baseline methods. For all methods, we use the default parameter settings.\footnote{For HCR, and ICP we use the R-packages from the references, for CISC, ACID, DR the corresponding Python code, and for IGCI, and RECI the Python package \code{causal discovery toolbox} \cite{KG19}.}

\subsection{Synthetic Data}

We generate synthetic data with ground truth $X \rightarrow Y$ using additive noise models $X:=N_X, Y:= f(X) + N_Y$ with $N_X \independent N_Y$ and multiplicative noise models $X:=N_X, Y := f(X)*N_Y$ with $N_X \independent N_Y$. We follow the data generation scheme as described in \cite{PJS11} for cyclic models. For the combinations $(b_x, b_y) \in \{(2,2),(3,3),(4,4),(5,5),(2,10),(10,2),(3,20),(20,3)\}$ we choose randomly a non constant function $f: \kX_X \rightarrow \kX_Y$ and independent distributions $P_X$ and $P_{N_Y}$. We sample $1000$ data points for each model and create $1000$ models for each range configuration $(b_x, b_y)$. Furthermore, we also simulate perfect interventions on $X$ by setting them to every value in $\kX_X$. Table \ref{tab:synthetic_results} shows the results for the data generated by an additive and multiplicative noise model, respectively. For the additive noise data, our method performs in a middle range compared to the alternatives and cannot compete with the leading methods. For multiplicative noise data IACM can compete with the leading alternative approaches for some configurations. The poor performance of IACM is because $\kM_{X \rightarrow Y}$ is too large and only encodes necessary conditions that a distribution originates from a causal model $X \rightarrow Y$. This is also supported by the fact that the overall performance of IACM+, were $\kM_{X \rightarrow Y}$ is further restricted based on some known identifiability results in the binary case, is around $4 \%$ better than that of IACM. We left a further structure refinement of $\kM_{X \rightarrow Y}$ and improvement of IACM for future research.

\begin{table}[htb]
\footnotesize
\begin{subtable}{\linewidth}
\begin{tabular}{|c||c|c|c|c|c|c|c|c|c|c|} \hline
$b_x, b_y$ & IACM & IACM+ & DR & IGCI & RECI & CISC & ACID & HCR & ICP \\ \hline
2,2 & 58,39,3 & 61,38,1 & 69,1,30 & 92,7,1 & 7,92,1 & 7,92,1 & 88,11,1 & 55,28,17 & {\bf 99},0,1 \\ \hline
3,3 & 54,44,2 & 45, 55,0 & 84,0,16 & 48,52,0 & 53,47,0 & 32,68,0 & {\bf 97},3,0 & 57,40,3 & 44,0,56 \\ \hline
4,4 & 49,33,18 & 62,38,0 & 89,0,11 & 48,52,0 & 51,49,0 & 44,56,0 & {\bf 98},2,0 & 60,39,1 & 50,0,50 \\ \hline
5,5 & 43,57,0 & 46,54,0 & 83,0,17 & 51,49,0 & 48,52,0 & 52,48,0 & {\bf 98},2,0 & 64,36,0 & 52,0,48 \\ \hline
2,10 & 48,42,10 & 59,41,0 & 88,0,12 & {\bf 100},0,0 & {\bf 100},0,0 & {\bf 100},0,0 & {\bf 100},0,0 & 59,36,5 & 66,0,34 \\ \hline
10,2 & 58,29,13 & 69,31,0 & {\bf 88},0,12 & 0,100,0 & 0,100,0 & 0,100,0 & 22,78,0 & 48,42,10 & 74,0,26 \\ \hline
3,20 & 57,41,2 & 31,69,0 & 61,0,39 & 100,0,0 & 99,1,0 & {\bf 100},0,0 & {\bf 100},0,0 & 61,38,1 & 78,0,22 \\ \hline
20,3 & 30,67,3 & 57,43,0 & 68,0,32 & 0,100,0 & 0,100,0 & 0,100,0 & 5,95,0 & 52,48,0 & {\bf 73},0,27 \\ \hline
 \end{tabular} 
 \subcaption{}
 \end{subtable}
 \begin{subtable}{\linewidth}
 \begin{tabular}{|c||c|c|c|c|c|c|c|c|c|c|} \hline
$b_x, b_y$ & IACM & IACM+ & DR & IGCI & RECI & CISC & ACID & HCR & ICP \\ \hline 
2,2  & 76,24,0 & 77,23,0 & 1,1,98 & 20,80,0 & 81,19,0 & 80,19,1 & 19,80,1 & 18,64,18 & {\bf 99},0,1  \\ \hline
3,3  & 63,36,2 & 51,49,0 & 4,2,94 & 36,64,0 & {\bf 65},35,0 & 56,44,0 & 51,49,0 & 42,48,10 & 23,0,77 \\ \hline
4,4  & 64,36,0 & {\bf 67},33,0 & 1,1,98 & 35,65,0 & 63,37,0 & 40,60,0 & 42,58,0 & 40,56,4  & 30,0,70 \\ \hline
5,5  & 47,51,2 & 45,55,0 & 1,0,99 & 52,48,0 & 51,49,0 & 34,66,0 & {\bf 59},41,0 & 52,46,2  & 35,0,65 \\ \hline
2,10 & 59,41,0 & 60,40,0 & 0,3,97 & 98,2,0  & {\bf 99},1,0  & {\bf 99},1,0  & 70,30,0 & 19,75,6  & 54,0,46 \\ \hline
10,2 & 67,33,0 & {\bf 68},32,0 & 3,1,96 & 0,100,0 & 25,75,0 & 5,95,0  & 9,91,0  & 54,28,18 & 66,0,34 \\ \hline
3,20 & 55,44,1 & 30,70,0 & 0,1,99 & {\bf 100},0,0 & 98,2,0  & {\bf 100},0,0 & 75,25,0 & 14,85,1  & 68,0,32 \\ \hline
20,3 & 55,45,0 & 59,41,0 & 1,0,99 & 0,100,0 & 9,91,0  & 0,99,1  & 2,98,0  & {\bf 63},31,6  & 50,0,50 \\ \hline
 \end{tabular} 
 \subcaption{}
 \end{subtable}
 \caption{\label{tab:synthetic_results}Results for synthetic data generated by additive (a) and multiplicative noise models (b) showing the percentage of correct identified directions, of wrong identified directions, and cases with no preferred directions. The highest percentage of correct identified directions is printed in {\bf bold}.}
  \normalsize
 \end{table}
 
%\begin{figure}[htb]
%\begin{minipage}{\linewidth}
%      \begin{minipage}{0.49\linewidth}
% \centering
%		  \includegraphics[scale=1.25]{result_linear_discrete.png}
%        \subcaption{}
%        \label{fig:discrete_linear}
% \end{minipage}
% \hfill
% \begin{minipage}{0.49\linewidth}
%\centering
%         \includegraphics[scale=1.25]{result_nonlin_discrete.png}
%        \subcaption{}
%        \label{fig:discrete_nonlinear}
%       \end{minipage}
%       \caption{Averaged accuracies of correct inferred causal direction for linear (a) and nonlinear (b) synthetic data relative to the difference in range size $|\kX_X| - |\kX_Y|$ for small range sizes.}
%       \label{fig.synthetic_results}
% \end{minipage}
%\end{figure}

\subsection{Real-World Data}

As real-world discrete data sets, we use $12$ cause-effect pairs from the acute inflammations dataset (Bladder) from \cite{DG19} as it has been used in \cite{PJS11}. Furthermore, we use $4$ anonymous discrete cause-effect pairs where food intolerances cause health issues (Food)\footnote{This dataset, given as discrete time-series data, has been provided by the author and the causal direction has been independently confirmed by medical tests.}, the Pittsburgh bridges dataset (Bridge, $4$ pairs) from \cite{DG19} as it has been used in \cite{CQZZH18}, and the Abalone dataset (Abalone, $3$ pairs) also from \cite{DG19}. The results in Table \ref{tab:real_world_results} show that IACM was able to reproduce a majority of the correct causal directions, whereas alternative approaches often failed in doing this.

\begin{table}[htb]
\footnotesize
\begin{tabular}{|c||c|c|c|c|c|c|c|c|c|c|} \hline
Dataset & IACM & IACM+ & DR & IGCI & RECI & CISC & ACID & HCR & ICP \\ \hline 
Bladder (12) & {\bf 9},1,2 & {\bf 9},1,2 & 3,0,9 & 6,3,3 & 2,8,2 & 1,8,3 & 5,4,3 &  6,2,4 & 7,1,4 \\ \hline
Abalone (3) & {\bf 3},0,0 & 2,1,0 & 0,0,3 &	{\bf 3},0,0 & {\bf 3},0,0 & {\bf 3},0,0 & {\bf 3},0,0 & {\bf 3},0,0 & 2,0,1 \\ \hline
Food	(4) & {\bf 4},0,0 & {\bf 4},0,0 &	0,1,3 & 1,3,0 & 1,3,0 & 1,3,0 & 0,4,0 & 2,1,1 & 2,0,2 \\ \hline
Bridge (4) & 3,1,0 & 2,2,0 & 0,1,3 & 1,3,0 & 3,1,0 & 3,1,0 & 3,1,0 & {\bf 4},0,0 & 2,0,2\\ \hline
 \end{tabular} 
 \\
 \caption{\label{tab:real_world_results}Results for real-world data showing the number of correct identified directions, of wrong identified directions, and cases with no preferred directions. The highest number of correct identified directions is printed in {\bf bold}.}
\normalsize
\end{table}

\section{Conclusions} \label{sec.discussion}

In this paper, we proposed a way how empirical distributions coming from observations and experiments can be approximated to ones that follow the restrictions enforced by an assumed causal model. This can be used to calculate probabilities of causation and leads to a new causal discovery method. In our experiments with synthetic data, we saw that the causal discovery method based on IACM is still too weak to compete with the current state-of-the-art methods. Nevertheless, for real-world datasets, IACM can compete with the leading alternative approaches. We suggest using more cautious causal discovery methods with a low false-positive rate like ICP to detect the causal direction and use IACM for calculation of causal probabilities to quantify the causal effect. The refinement and improvement of causal discovery based on IACM are left for future research.

%
%\section*{Broader Impact}
%
%As all methods for causal discovery that use observational and/or data from implicit interventions the work in this paper could help to avoid unethical experiments. Furthermore, it contributes to a more relieable detection of causal relations, since it fills a gap in the existing causal discovery landscape. Therefore, our research can help during the evaluation and design of studies with few discrete features and contribute to more solid conclusions of those studies. On the other hand there is a potential risk that our method is used for data that are not following the assumptions of this paper. This may lead to wrong causal directions and to wrong conclusions, but can be avoided by checking the assumptions on the data before applying our method. Finally, it should be noted that this article may inspire future research projects in the field.

\bibliographystyle{myamsalpha}
\bibliography{mrabbrev, references}

\section*{Supplementary Material for Information-Theoretic Approximation to Causal Models}

\title{Supplementary Material for Information-Theoretic Approximation to Causal Models}

\maketitle

\appendix
\section{Proofs}

\subsection{Proof of Lemma 1}
\begin{lemma}{1}
The set of joint probability distributions for $V$ which fulfil condition (1) is given as
\begin{eqnarray*}
\kM_{X \rightarrow Y} & = & \left \{P \in \kP(\kX_V) \mid \pi_{X, Y, Y_{a}} P(a, y_a, \overline{y}_{a}) = 0 \right. \\
& & \qquad \qquad \qquad \left. \forall \; y_a \in \kX_{Y_a}, \overline{y}_a \in \kX_{Y_a} \backslash \{y_a\}, a \in \kX_X \right \}.
\end{eqnarray*}
\end{lemma}
\begin{Proof}
The consistency condition (1) implies the following relation for some $P \in \kP(\kX_V)$ and $a\in \kX_X$
\begin{eqnarray*}
\pi_{X, Y} P(a, y_a) & = & \pi_{X, Y, Y_{a}} P(a, y_a, y_{a}).
\end{eqnarray*}
These relation implies
\begin{eqnarray*}
\pi_{X, Y, Y_{a}} P(a, y_a, \overline{y}_{a}) = 0, {\rm \;for}\;a \in\kX_X,
\end{eqnarray*}
which characterizes the joint distributions that satisfy (1).
\end{Proof}

\subsection{Proof of Proposition 1}

\begin{satz}{1}
The optimization problem (3) simplifies to the following linear optimization problem
$$
\max_{\substack{\hat P \in \kP(\kX_V),\\ \pi_{X,Y} \hat P = P_{XY}, \pi_{Y_{a}} \hat P = P_{Y_{a}},a\in \kX_X}} S_{X \rightarrow Y}(\hat P),
$$
with $S_{X \rightarrow Y}(\hat P):= \sum_{z \in \supp(\kM_{X \rightarrow Y})} \hat P(z)$.
\end{satz}
\begin{Proof}
We first consider the inner minimization problem of (3) for a given joint distribution $\hat P \in \kP(\kX_V)$. This is a constrained optimization problem where the constraints in $\kM_{X \rightarrow Y}$ are equivalent to the equation
$$
S_{X \rightarrow Y}(\tilde P) = 1,
$$
since $\tilde P$ is a probability distribution. Therefore, the Lagrange functional of this minimization problem reads
$$
\Lambda(\tilde P) := D(\tilde P || \hat P) + \lambda \left (S_{X \rightarrow Y}(\tilde P) - 1
\right ),
$$
with $\lambda$ as Lagrange multiplier. Using the Lagrange multiplier method we obtain explicit expressions for the approximating distribution $\tilde P \in \kM_{X \rightarrow Y}$
$$
\tilde P (z)= \frac{\hat P(z)}{S_{X \rightarrow Y}(\hat P)}
$$
for $z \in \supp(\kM_{X \rightarrow Y})$ and $\tilde P(z) = 0$ for all $z \notin \supp(\kM_{X \rightarrow Y})$. Thus we have solved the inner minimization problem explicitly and the relative entropy simplifies to
$$
D( \tilde P || \hat P) = - \log(S_{X \rightarrow Y}( \hat P)).
$$
Therefore, we can now optimize on the space of possible joint distributions and (3) simplifies to
$$
\max_{\substack{\hat P \in \kP(\kX_V),\\ \pi_{X,Y}\hat P = P_{XY}, \pi_{Y_{a}} \hat P = P_{Y_{a}},a\in \kX_X}} \log(S_{X \rightarrow Y}(\hat P)).
$$
Since $\log$ is a monotone function it suffices to maximize $S_{X \rightarrow Y}(\hat P)$ given the constraints. But this is nothing than a linear optimization problem which can be solved by linear programming using the simplex algorithm, see, for example, \cite{CLRS01}.
\end{Proof}

\section{Application to Timeseries Data}

Algorithm 1 can also be applied when we assume that the underlying causal model has a time lag of $T$, which is $X_{t-T} \rightarrow Y_t$, and the observational and interventional data have a time order. We only have to shift the incoming data for $X_t$ and $Y_t$ so that Algorithm 1 applies to $X_{t}, Y_{t+T}$, and has to take care that we preserve the data order during preprocessing steps. If we do not know the exact time lag we can run the approximation several times with different time lags to find
the approximation with the lowest error. 

\section{Extensions to trivariate causal models}

In the following we list the support set characterizations for some causal models with three variables.

Causal model: $X \leftarrow Z \rightarrow Y$ with intervention on $Z$,\\
$V = \{X,Y,Z\} \cup \bigcup_{a \in \kX_Z} \{X_a, Y_a\}$, \\
\begin{eqnarray*}
M_{X \leftarrow Z \rightarrow Y} \! \!\!\!\! & = &\!\!\!\!\!  \left \{P \in \kP(\kX_V) \mid \pi_{X,Y,Z,X_a, Y_a} P(x_a,y_a, a, \overline{x}_a, y_a) = \right. \\
& & \; \quad \qquad \qquad \left. \pi_{X,Y,Z,X_a, Y_a} P(x_a,y_a, a, x_a, \overline{y}_a) = 0 \right. \\
& & \; \quad \qquad \qquad \left. \forall x_a \in \kX_{X_a}, y_a \in \kX_{Y_a}, \overline{x}_a \in \kX_{X_a} \backslash \{x_a\}, \overline{y}_a \in \kX_{Y_a} \backslash \{y_a\}, a \in \kX_Z \right \}.
\end{eqnarray*}  
If we cannot observe $Z$ but know the range $\kX_Z$ and know (or can reconstruct) $X_a, Y_a$ for $a \in \kX_Z$ we can also state a support set characterization of condition (1) for such a model. \\
Causal model: $X \leftarrow (Z) \rightarrow Y$ with unobserved intervention on $Z$,\\
$V = \{X,Y\} \cup \bigcup_{a \in \kX_Z} \{X_a, Y_a\}$, \\
\begin{eqnarray*}
M_{X \leftarrow (Z) \rightarrow Y} \! \!\!\!\! & = &\!\!\!\!\!  \left \{P \in \kP(\kX_V) \mid \pi_{X,Y,X_a, Y_a} P(x_a,y_a, \overline{x}_a, y_a) = \pi_{X,Y,X_a, Y_a} P(x_a,y_a, x_a, \overline{y}_a) = 0 \right. \\
& & \; \quad \qquad \qquad \left. \forall x_a \in \kX_{X_a}, y_a \in \kX_{Y_a}, \overline{x}_a \in \kX_{X_a} \backslash \{x_a\}, \overline{y}_a \in \kX_{Y_a} \backslash \{y_a\}, a \in \kX_Z \right \}.
\end{eqnarray*}  

Causal model: $Z \rightarrow X \rightarrow Y$ with interventions on $Z$ and $X$,\\
$V = \{X,Y,Z\} \cup \bigcup_{a \in \kX_Z} \{X_a\} \cup \bigcup_{b \in \kX_X}, \{Y_b\}$, \\
\begin{eqnarray*}
M_{Z \rightarrow X \rightarrow Y} \! \!\!\!\! & = &\!\!\!\!\!  \left \{P \in \kP(\kX_V) \mid \pi_{X,Y,Z,X_a, Y_{x_a}} P(x_a,y_{x_a},a, \overline{x}_a, y_{x_a}) = \right. \\
& & \; \quad \qquad \qquad \left. \pi_{X,Y,Z,X_a, Y_{x_a}} P(x_a,y_{x_a},a, x_a, \overline{y}_{x_a}) = 0 \right. \\
& & \; \quad \qquad \qquad \left. \forall y_{x_a} \in \kX_{Y_{x_a}}, \overline{y}_{x_a} \in \kX_{Y_{x_a}} \backslash \{y_{x_a}\}, \overline{x}_a \in \kX_{X_a} \backslash \{x_a\}, x_a \in \kX_{X_a}, a \in \kX_Z \right \}.
\end{eqnarray*}  

Causal model with hidden variable $Z$: $(Z) \rightarrow X \rightarrow Y$ with unobserved interventions on $Z$ but observed interventions on $X$,\\
$V = \{X,Y\} \cup \bigcup_{a \in \kX_Z} \{X_a\} \cup \bigcup_{b \in \kX_X}, \{Y_b\}$, \\
\begin{eqnarray*}
M_{(Z) \rightarrow X \rightarrow Y} \! \!\!\!\! & = &\!\!\!\!\!  \left \{P \in \kP(\kX_V) \mid \pi_{X,Y,X_a, Y_{x_a}} P(x_a,y_{x_a},\overline{x}_a, y_{x_a}) = \right. \\
& & \; \quad \qquad \qquad \left. \pi_{X,Y,X_a, Y_{x_a}} P(x_a,y_{x_a},x_a, \overline{y}_{x_a}) = 0 \right. \\
& & \; \quad \qquad \qquad \left. \forall y_{x_a} \in \kX_{Y_{x_a}}, \overline{y}_{x_a} \in \kX_{Y_{x_a}} \backslash \{y_{x_a}\}, \overline{x}_a \in \kX_{X_a} \backslash \{x_a\}, x_a \in \kX_{X_a}, a \in \kX_Z \right \}.
\end{eqnarray*}  

Causal model: $Z \rightarrow Y \leftarrow X$ with interventions on $Z$ and $X$,\\
$V = \{X,Y,Z\} \cup \bigcup_{a \in \kX_Z} \{Y_a\} \cup \bigcup_{b \in \kX_X}, \{Y_b\}$, \\
\begin{eqnarray*}
M_{Z \rightarrow Y \leftarrow X} \! \!\!\!\! & = &\!\!\!\!\!  \left \{P \in \kP(\kX_V) \mid \pi_{X,Y,Z,Y_a, Y_b} P(b,y_a,a, y_a, \overline{y}_a) =  \right. \\
& & \; \quad \qquad \qquad \left. \pi_{X,Y,Z,Y_a, Y_b} P(b,y_a,a, \overline{y}_a, y_a) = \right. \\
& & \; \quad \qquad \qquad \left. \pi_{X,Y,Z,Y_a, Y_b} P(b,y_a,a, \overline{y}_a, \overline{y}_a) = 0 \right. \\
& & \; \quad \qquad \qquad \left. \forall y_a \in \kX_{Y_a}, \overline{y}_a \in \kX_{Y_a} \backslash \{y_a\}, b \in \kX_X, a \in \kX_Z \right \}.
\end{eqnarray*}  

Causal model with hidden variable $Z$: $(Z) \rightarrow Y \leftarrow X$ with unobserved interventions on $Z$ and observed interventions on $X$,\\
$V = \{X,Y\} \cup \bigcup_{a \in \kX_Z} \{Y_a\} \cup \bigcup_{b \in \kX_X}, \{Y_b\}$, \\
\begin{eqnarray*}
M_{(Z) \rightarrow Y \leftarrow X} \! \!\!\!\! & = &\!\!\!\!\!  \left \{P \in \kP(\kX_V) \mid \pi_{X,Y,Y_a, Y_b} P(b,y_a, y_a, \overline{y}_a) =  \right. \\
& & \; \quad \qquad \qquad \left. \pi_{X,Y,Y_a, Y_b} P(b,y_a, \overline{y}_a, y_a) = \right. \\
& & \; \quad \qquad \qquad \left. \pi_{X,Y,Y_a, Y_b} P(b,y_a, \overline{y}_a, \overline{y}_a) = 0 \right. \\
& & \; \quad \qquad \qquad \left. \forall y_a \in \kX_{Y_a}, \overline{y}_a \in \kX_{Y_a} \backslash \{y_a\}, b \in \kX_X, a \in \kX_Z \right \}.
\end{eqnarray*}  

All those extensions lead to different objective functions $S_C$ that can be maximized (with adapted constraints) as described in Algorithm 1. We left it for future research to investigate to what extend the models with a confounding variable can be used as a confounder detector.

\section{Causal Discovery with Additive Noise Constraints}

Based on the characterization of reversible and irreversible additive noise models (ANMs) for binary random variables $X$ and $Y$ in \cite{PJS11} we can further restrict the set $\kM_{X \rightarrow Y}$ such that distributions from reversible models are more likely to be avoided. From Section III C. in \cite{PJS11} we know that  \begin{equation}
\begin{aligned} \label{anm_constraint}
P_{X,Y}(X=0,Y=0) & = P_{X,Y}(X=1,Y=1), \quad {\rm or}  \\
P_{X,Y}(X=0,Y=1) & =  P_{X,Y}(X=1,Y=0)
\end{aligned}
\end{equation}
leads to a reversible ANM. That means if there is an ANM from $X$ to $Y$, then there is also an ANM from $Y$ to $X$ given that (\ref{anm_constraint}) is fulfilled. We use this to add a penalty term to $S_{X \rightarrow Y}$ such that (\ref{anm_constraint}) will be avoided in the optimization procedure. We define \mbox{$\delta_1:= |P_{X,Y}(X=0,Y=0) - P_{X,Y}(X=1,Y=1)|$} and \mbox{$\delta_2:=|P_{X,Y}(X=0,Y=1) -  P_{X,Y}(X=1,Y=0)|$} and get
$$
S_{1,X \rightarrow Y} = S_{X \rightarrow Y} + \delta_1, \qquad S_{2,X \rightarrow Y} = S_{X \rightarrow Y} + \delta_2.
$$
When maximizing $S_{i, X\rightarrow Y}$ instead of $S_{X \rightarrow Y}$ we also maximize $\delta_i$ for $i \in \{1,2\}$. Resolving the absolute value and simplifying we end up in $4$ objective functions we can maximize:
\begin{equation}
\begin{aligned}  \label{s_functions}
S_1 & = 2 P_{0000} + 2 P_{0001} + P_{0110} + P_{0111} + P_{1000} + P_{1010}, \\
S_2 & = P_{0110} +P_{0111} + P_{1000} + P_{1010} + 2P_{1101} + 2P_{1111}, \\
S_3 &= P_{0000} + P_{0001} + 2P_{0110} + 2P_{0111} + P_{1101} + P_{1111}, \\
S_4 &= P_{0000} + P_{0001} + 2P_{1000} + 2P_{1010} + P_{1101} + P_{1111}.
\end{aligned}
\end{equation}
These objective functions for the optimization procedure contain now some sort of asymmetry and exclude unfavorable distributions. We integrate this into the existing IACM algorithm by maximizing each of the functions in (\ref{s_functions}) and chose the model approximation with the lowest approximation error. We call that version IACM+. Note that condition (\ref{anm_constraint}) is not sufficient for reversible ANMs. The other conditions stated in \cite{PJS11} cannot be integrated into our procedure in such a straightforward way. However, in the experiments, we saw that IACM+ leads to slightly better results than IACM for additive and multiplicative noise data.

\section{Experiments}
\subsection{Data Preprocessing}

Algorithm 3 includes several preprocessing steps for sample data from $X$ and $Y$ that can be parametrized. The preprocessing steps split the data into observational data and interventional data for $X$ and $Y$ accordingly if they are not given in explicit form. The current implementation supports the following options:
\begin{itemize}
	\item \code{none}: split the data for $X$ and $Y$ in the middle of the data sets into observational and interventional data, accordingly.
	\item \code{split}: filter $X, Y$ by each intervention on $X$ and draw without replacement from each filtered subset data points for observational and interventional data to obtain equally sized sets for observations and interventions with similar variance.
	\item \code{split-and-balance}: works like \code{split} but draws with replacement such that each intervention subset has equal size and data from interventions with fewer samples are lifted. This method tries to balance out imbalanced intervention subsets.
\end{itemize}

\subsection{Parameter Setting in Experiments}

In the experimental runs we use the following parameter configuration for IACM and IACM+. For all datasets we use \code{none} as preprocessing and the local approximation error $D(\pi_{X,Y} \tilde P || \pi_{X,Y} \hat P)$ as decision metric. If the ranges of $X$ or $Y$ are binary we use $b_x = b_y=2$ and $b_x = b_y = 3$ otherwise.

\end{document}